\def\year{2021}\relax
\documentclass[letterpaper]{article} 
\usepackage{aaai21}  
\usepackage{times}  
\usepackage{helvet} 
\usepackage{courier}  
\usepackage[hyphens]{url}  
\usepackage{graphicx} 
\def\UrlFont{\rm}  
\usepackage{natbib}  
\usepackage{caption} 
\usepackage{makecell}
\usepackage{amsmath,mathtools}
\usepackage{adjustbox}
\usepackage[ruled,vlined,linesnumbered]{algorithm2e}
\usepackage{comment}
\usepackage{bm}
\usepackage{amsfonts}
\usepackage{subcaption}
\usepackage{multirow}
\nocopyright

\title{Weight Equalizing Shift Scaler-Coupled Post-training Quantization}
\author{
Jihun Oh, 
SangJeong Lee, 
Meejeong Park, 
Pooni Walagaurav, 
Kiseok Kwon \\
}
\affiliations{
Samsung Research, Samsung Electronics Co. \\
33, Seongchon-gil, Seocho-gu, Seoul, 06765, Republic of Korea \\
\{jihun2331.oh, sj94.lee, mia.park, pt.gaurav, kiseok.kwon\}@samsung.com    
}

\begin{document}

\maketitle

\begin{abstract}
Post-training, layer-wise quantization is preferable because it is free from retraining and is hardware-friendly. Nevertheless, accuracy degradation has occurred when a neural network model has a big difference of per-out-channel weight ranges. In particular, the MobileNet family has a tragedy drop in top-1 accuracy from 70.60\% $\sim$ 71.87\% to 0.1\% on the ImageNet dataset after 8-bit weight quantization. To mitigate this significant accuracy reduction, we propose a new weight equalizing shift scaler, i.e. rescaling the weight range per channel by a 4-bit binary shift, prior to a layer-wise quantization. To recover the original output range, inverse binary shifting is efficiently fused to the existing per-layer scale compounding in the fixed-computing convolutional operator of the custom neural processing unit. The binary shift is a key feature of our algorithm, which significantly improved the accuracy performance without impeding the memory footprint. As a result, our proposed method achieved a top-1 accuracy of 69.78\% $\sim$ 70.96\% in MobileNets and showed robust performance in varying network models and tasks, which is competitive to channel-wise quantization results.
\end{abstract}


\section{Introduction}

Quantization involves mapping continuous real values to discrete integers, which is demanding to reduce the computational and memory costs. Inference using quantized weights and feature maps in fixed-precision (e.g., 8 bits) can compute many operations per second and reduce memory bandwidth and power consumption. In detail, the chip area of a multiplier is proportional to the square of precision so low precision-based computing can lead to either a reduction in chip area or an increase in computing capability under the same chip area. Also, the energy consumption of 8-bit multiplier and adder is 18 x and 30 x lower than 32-bit floating-point, respectively \cite{horowitz20141}. Therefore, quantization is necessary for deep learning models to run on energy-efficient specialized hardware such as neural processing units (NPU) \cite{chen2016eyeriss,judd2016stripes,albericio2017bit,jouppi2017datacenter,lee2018unpu,zhao2019linear}.

According to the intervention point of quantization, there are two forms: quantization-aware training (QAT) and post-training quantization (PTQ). In detail, QAT \cite{han2015deep,courbariaux2015binaryconnect,hubara2016binarized,lee2018deeptwist,jung2019learning} is to simulate quantization during the time of training from scratch or fine-tuning, enabling to apply the effect of quantization simultaneously. Training can compensate for accuracy degradation due to quantization so we can reach even lower precision such as 4 bits. However, QAT is not always available in real practice because it requires large training or validation dataset with label information and complicated training settings, and takes a lot of time to get results. 

Alternatively, PTQ \cite{krishnamoorthi2018quantizing,nagel2019data,banner2019post,zhao2019improving,cai2020zeroq} can quantize pre-trained models in a single shot and is free from large training and validation datasets, but just small representative dataset without labels for activation quantization. It is simple and fast because of unnecessity of retraining. Nevertheless, quantizing full precision models into fixed point without retraining in PTQ yields quantization error on weights because of the limitation of the representable number of bits. Furthermore, as shown in Fig. 1, there are often significant differences in weight ranges across out channels, and relatively narrow-ranged weights in specific channels are quantized to just a few quantized values. This can deteriorate the accuracy in the layer-wise quantization (LWQ) scheme. To address this problem, per-channel bit allocation \cite{banner2019post} and ZeroQ \cite{cai2020zeroq} were introduced but mixed precision is more complicated to implement in hardware than homogeneous precision. Most commodity hardwares do not support efficient mixed precision computation due to chip area constraints \cite{liu2020post}. Outlier-channel splitting \cite{zhao2019improving} minimized quantization error by halving weights belonging to outliers in the distribution and increasing channels. However, this can increase the cost of network size overhead. 

In this work, we focus on proposing a straightforward but powerful manner to recover the baseline accuracy in the PTQ scheme.

$\bullet$ To minimize the accuracy drop after PTQ, a shift-based channel-wise weight scaling method is introduced prior to the quantization process. This aligns weight ranges across channels approximately so quantization error is drastically reduced. Inverse binary shifting is efficiently fused to the existing per-layer scale compounding to recover the original output range. 

$\bullet$ The proposed work is validated using state-of-the-art models in varying tasks (image classification, object detection, super-resolution). In particular, we use a large scale dataset such as ImageNet and focus on up-to-date models considering both accuracy and efficiency rather than traditional deep models with high redundancy. Plus, we intactly utilize only open pre-trained models without retraining.

\section{Previous Works}

There are a couple of comparable studies to reduce quantization error using a homogeneous precision level in PTQ. To name representative methods, there are channel-wise quantization (CWQ) \cite{krishnamoorthi2018quantizing} and data-free quantization (DFQ) \cite{nagel2019data}.
\subsection{Pitfalls of Channel-wise Quantization}
Layer-wise quantization produces a single set of scale compound and zero point per layer, thus unabling to preserve weight values in the relatively narrow-ranged channels after quantization. \\
\textbf{1. Bigger parameter size:} To address the problem, channel-wise quantization gives as many sets of the parameters as the number of channels. Despite a guarantee of high accuracy, however, the bigger parameter size can be burdensome if deploying the model on edge devices with limited memory resource. That is because it can cause a latency issue due to a delayed loading of the parameters increased per each channel.\\
\textbf{2. Bias overflow:} In addition, the channel-wise scales, $(max_i-min_i)/(2^{bits}-1)$, in depth-wise convolution are sometimes quite small so that as quantizing the bias by the multiplication of the scale of input and scale of weight, $B^i/(s_{in} s_w^i)$, it incurs a 32-bit overflow issue in conventional NPU. To prohibit the overflow, the $[min, max]$ range determining the scale should be forced to be adjusted. That is, if the computed scale is less than the low bound (e.g., 1e-5), the max range is extended, fitting to the scale bound. Though, this ad hoc recipe discards the precision gain of channel-wise quantization. 
\subsection{Pitfalls of Data-free Quantization}
To tackle the accuracy drop in the MobileNet family, data-free quantization \cite{nagel2019data} rescales the weight scales across subsequent layers by making use of a scale-equivariance property of activation functions and the scale factor is searched by an iterative optimization. \\
\textbf{1. Model modification:} However, ReLUN(N=6) requires a different cut off per channel after applying the scale equlization procedure. Due to hardware complexity, this algorithm supports regular ReLU activation function efficiently. Thus, it requires the replacement of the activation function. \\
\textbf{2. Retraining:} After the architecture change, task accuracy could decrease because of the change of distribution after activation function. In the reference paper, the authors claim little accuracy drop in PyTorch but we experienced nontrivial accuracy degradation ($70.60\%\rightarrow55.45\%$, $71.87\%\rightarrow64.90\%$) for MobileNet v1/v2 \footnote[1]{https://github.com/fchollet/deep-learning-models/releases} pretrained in Tensorflow Keras. This can depend on the training quality of pre-trained models. To recover the baseline accuracy, further retraining of the changed model using a train dataset is required, however, which is undesirable in post-training quantization.   

\section{Method}

\begin{figure}[htp]
    \centering
    \includegraphics[width=8cm]{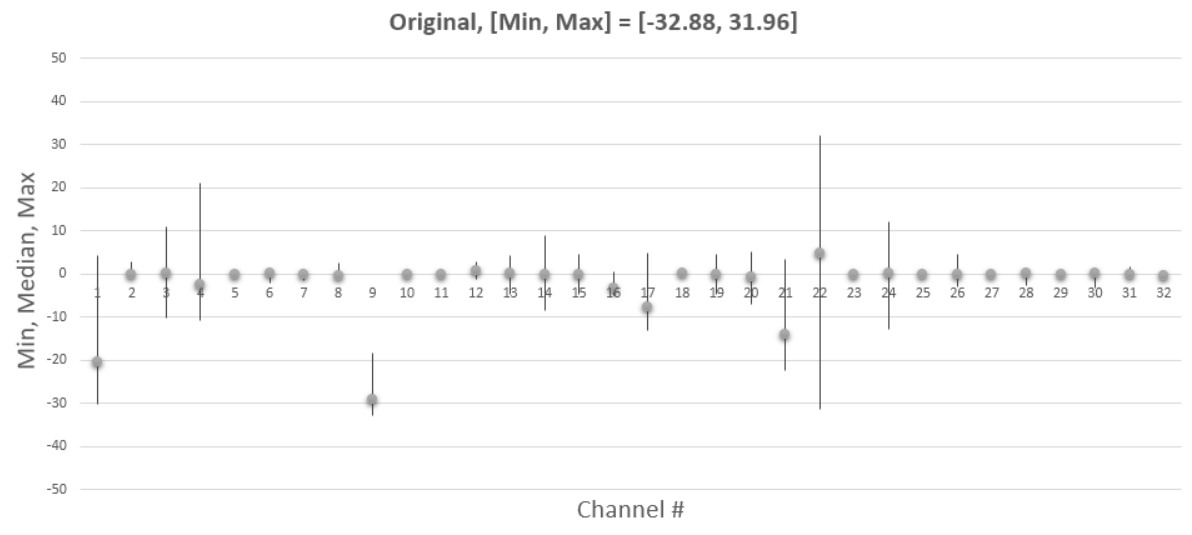} \\
    \includegraphics[width=8cm]{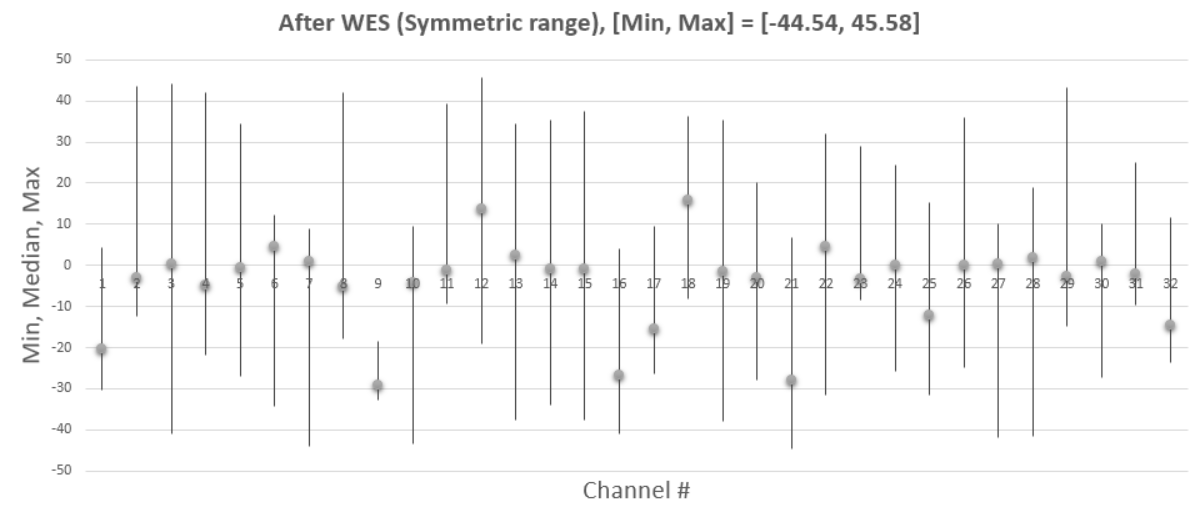}
    \caption{Channel-wise weight ranges and rescaled weight ranges by WES of the 1st depthwise convolutional layer in MobileNet v1. Min, median, and max values are plotted.}
    \label{fig:mv1_dist_before_after_wes}
\end{figure}

\subsection{Fundamental of Uniform Affine Quantization}
For each layer, quantization is parameterized by the
number of quantization levels ($2^{bit}$) and a clamping range ($a, b$), and is
performed by applying the element-wise quantization function
$q_w$ defined as follows \cite{jacob2018quantization}:

\begin{equation}
 \text{clamp}[x; a, b] = \text{min}(\text{max}(x, a) ,b),
\end{equation}
\begin{equation}
  s = \frac{max-min}{2^{bits}-1}, \quad z = \lfloor \frac{-min}{s_w} \rceil,
\end{equation}
\begin{equation}
  q = \lfloor \frac{1}{s}(\text{clamp}[w; min, max]-min) \rceil,
\end{equation}
where the boundaries $[min, max]$ are nudged by small amounts so that
value 0.0 is exactly representable as an integer zero point, $z$, after quantization: $min \leftarrow s*\lfloor min/s \rceil, \quad max \leftarrow max + s*\lfloor min/s \rceil - min$. The function $\lfloor \cdot \rceil$ rounds to the nearest integer. The scale, $s$, indicates the step size of quantization. Uniform affine quantization is also called asymmetric quantization because the quantized range is fully utilized. We exactly map the min/max values from the float range to the min/max of the quantized range due to the existence of the zero point.

\subsection{Weight Equalizing Shift Scaler}
The proposed method introduces a new weight equalizing shift scaler (WES)-coupled post training quantization having a new quantized parameter called “channel-wise shift scale” and its inverse scaling’s fusion to a convolutional operator in NPU.

\begin{figure*}
    \includegraphics[width=16cm]{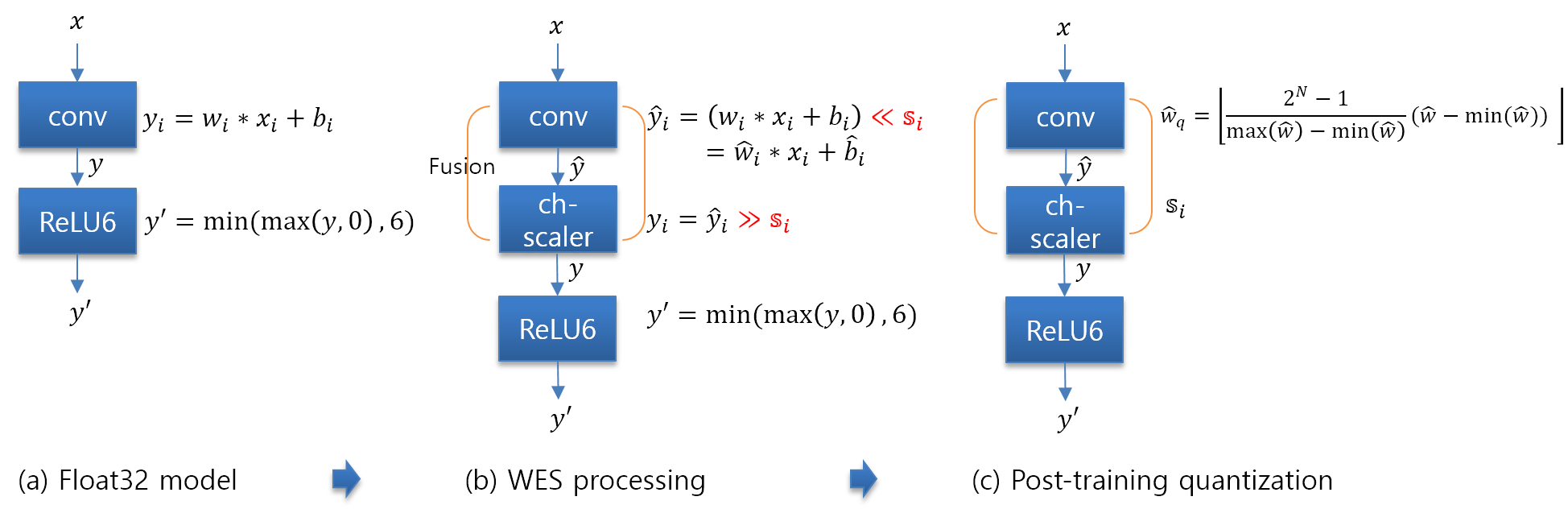}
    \caption{The WES-coupled post-training quantization procedure. “Channel-wise shift scale” and its inverse scaling’s fusion to a convolutional operator in the neural processing unit (NPU).}
    \label{fig:wes_diagram}
\end{figure*}

\subsubsection{Initialization of a total range}

Rescaling the range of original weights to the optimal range per channel before quantization is required to minimize the quantization error. To do so, convolutional weights and bias first fold its following batch norm (BN) layer (Fig. \ref{fig:wes_diagram} (a)) since most NPUs basically support BN-folded convolutional operation for speed-up inference. 
An original range $r_{i}$ is defined as a symmetric range, which is a maximum between absolute values of max and min of weights in each channel $i$. A total range $\hat{r}$ is the target range in WES to which all channel-wise ranges in a layer will be fitted. This is initialized as a maximum value over all $r_{i}$ in a layer:
\begin{equation}
r_{i}=2 * \text{max}(|max_{i}|, |min_{i}|), \quad
\hat{r}=\text{max}(r_{i}).
\end{equation}

An integer format of a channel-wise shift scale factor $\mathbb{S}_{i}$ to rescale the original range for $i$-th output channel tensor into the total range is
\begin{equation}
\mathbb{S}_i=\lfloor log_2(\hat{r}/r_{i}) \rfloor,
\end{equation}
where a floor function $\lfloor \cdot \rfloor$ gives the greatest integer less than or equal to input.

\subsubsection{Iterative search for an optimal total range}
For more minute investigation of the optimal total range $\hat{r}$, we use a fake quantization (quantization followed by dequantization)-based optimization scheme where quantization error is defined as a cost function to be minimized. \\
\indent The WES-coupled fake quantization process is shown in Algorithm \ref{alg:WES}. In detail, weights are first channel-wise shifted by $s_i$ (Line 7 in Algorithm 1), followed by uniform affine layer-wise quantization (FP32$\rightarrow$UINT8) and dequantization (UINT8$\rightarrow$FP32) process (Line 8-10) as emulating inference-time quantization, and then channel-wise shifted in the reversed direction (Line 11). Lastly, the quantization error can be measured in $l2$-norm between original weights and WES-coupled fake quantized weights so that bigger quantization error can be more penalized (Line 12). Note that the floor and round functions used in the fake quantization process have discontinuities at the integers, for which derivatives may not be known. Thus, we use the heuristic non-linear optimization methods such as Nelder–Mead \cite{nelder1965simplex} or Bayesian optimization  \cite{pelikan1999boa} where they iteratively search for the local minima of the cost function while the absolute error between iterations is greater than a tolerance.

\subsubsection{Channel-wise shift scale}
Given the optimal channel-wise shift scale $\mathbb{S}_i$, the original weights $w_{i}$ and bias $b_{i}$ are rescaled to $\hat{w}_{i}$ and $\hat{b}_{i}$ by a shift operation (Line 14 in Algorithm 1):
\begin{equation}
\hat{w}_i = w_i \ll \mathbb{S}_i, \quad
\hat{b}_i = b_i \ll \mathbb{S}_i,
\end{equation}

This WES process shifts the weights closely fitting to the total weight range (Fig. \ref{fig:wes_diagram} (b)). Next, a scale ($s_w$) and a zero point ($z_w$) are derived by uniform affine layer-wise quantization (Fig. \ref{fig:wes_diagram} (c)). The scale specifies the step size of the quantizer and the zero point is an integer mapped to the floating point zero value \cite{krishnamoorthi2018quantizing}.
In particular, the distribution of the channel with a relatively narrow range is stretched out so it comes to having low quantization error after WES-coupled quantization. The process so far is performed in compile time and a resultant quantized model is saved to a binary file.   

\begin{algorithm}
\SetAlgoLined
 $w, b \leftarrow $ BN\_fold $(w, b, \gamma, \beta, \mu, \sigma^2)$ \\
 $[min_{i}, max_{i}] = [\min(w_i), \max(w_i)]$ \\
 $r_{i}=2*\max(|max_{i}|, |min_{i}|)$ \\
 $\hat{r}=\max(r_{i})$ // initialization of trainable variable \\ 
 \While{$|\phi - \phi_{old}| > tol $}{
  $\mathbb{S}_i=\lfloor log_2(\hat{r}/r_{i}) \rfloor$ \\
  $\hat{w}_i = w_i \ll \mathbb{S}_i$ \\
  $[min, max] = [\min(\hat{w}), \max(\hat{w})]$ \\
  $s_w = \frac{max-min}{2^{bits}-1}$ \\
  $\hat{w}_i^* = s_w * \lfloor \frac{1}{s_w}(\hat{w}_i - min)\rceil + min$ \\
  $w_i^* = \hat{w}_i^* \gg \mathbb{S}_i$ \\
  $\phi = \frac{1}{n} \sum{(w_i - w_i^*)^2}$ \\
 }
 $\hat{w}_i = w_i \ll \mathbb{S}_i \quad \hat{b}_i = b_i \ll \mathbb{S}_i$
 \caption{Weight equalizing shift scaler (WES)}
 \label{alg:WES}
\end{algorithm}

\subsection{WES-coupled Fixed-point Computing in NPU}
In this section, we describe how to implement a convolutional operation, adapting the change by our proposed WES to the fixed-computing NPU. From end to end of network, all the computations are performed in the integer format. 

The rescaled output feature maps are resulted from matrix multiplication and vector addition using weights and bias rescaled by WES described in the previous section. Their original output range needs to be recovered before an activation function. To do so, it should be followed by a channel-wise inverse shift scaling, $s_i^{-1}$ (Fig. \ref{fig:wes_diagram} (b)). However, it is undesirable to have an additional layer that did not exist in the original model.

To fuse the convolutional layer and the channel-wise inverse shift scaling layer, it is necessary to design a WES-specialized convolutional operator in NPU as shown in Algorithm \ref{alg:fixed-computing}. In order to produce a fixed-point output $q_{out}$ using given a fixed-point input $q_{in}$, the NPU operators take supported integer-format weights $q_w$ (uint8), and its quantization-related parameters such as scales (float32), $s_w$, $s_{in}$, $s_{out}$ and zero points (uint8), $z_w$, $z_{in}$, $z_{out}$, respectively. There exists a single set of those quantization parameters per each layer in layer-wise quantization.
In the fixed-computing convolutional operation \cite{jacob2018quantization}, the output $q_{out}^i$ for channel $i$ is computed by 
\begin{equation}
\begin{split}
&s_{out}(q^i_{out}-z_{out}) \\
&=\sigma\left(\sum_j \sum_{h_w, w_w}s_w(q^{ij}_w - z_w)s_{in}(q^j_{in} - z_{in}) + B^i\right).
\end{split}
\end{equation}
This can be refactored to
\begin{equation}
\begin{split}
&q^i_{out}=z_{out} + \\ 
&\sigma\left(\frac{s_{in}s_w}{s_{out}}\left( \left( \sum_j \sum_{h_w, w_w} (q^{ij}_w - z_w) (q^j_{in} - z_{in}) \right) + q^i_B \right) \right), \\
\end{split}
\end{equation}
where $\sigma(\cdot)$ is a piece-wise linear activation function (e.g., ReLU, ReLUN) and $q_B^i$ denotes the quantized bias through $B^i/(s_{in} s_w)$. $h_w$ and $w_w$ indicate height and width of a weight tensor, and $i$ and $j$ denote indices of output and input channels.
A scale compound $(s_{in}s_w)/s_{out}$ is the same over the channels, and is approximated to a binary floating-point number format with a fixed number of significant digits (mantissa) and scaled using an exponent: $M*2^{s}$. $M$ is a 32-bit mantissa representing a value in the interval $[0.5,1)$ and $s$ is a 6-bit integer-format exponent, enabling a shift in the left or right direction. 
\begin{table}
\caption{Weight parameter specification after layer-wise vs. channel-wise vs. WES-coupled layer-wise quantization. $M$ and $s$ are the mantissa and the exponent of a scale compound and the subscript $i$ denotes a channel index out of $N$ channels. $\mathbb{S}_i$ is a channel-wise shift scale. DFQ has the same spec. as LWQ.}
\label{tab:lwq_cwq_wes}
\centering
\begin{tabular}{p{0.3\linewidth}p{0.05\linewidth}p{0.10\linewidth}p{0.12\linewidth}p{0.12\linewidth}}
\hline
Weight params & bits & LWQ & CWQ & WES \\
\hline
\hline
\multirow{3}{*}{Scale compound}
         & \multicolumn{1}{l}{32} & \multicolumn{1}{l}{$M$} & \multicolumn{1}{l}{$M_i \times N$} & \multicolumn{1}{l}{$M$} \\
         & \multicolumn{1}{l}{6} & \multicolumn{1}{l}{$s$} & \multicolumn{1}{l}{$s_i \times N$} & \multicolumn{1}{l}{$s$} \\
         & \multicolumn{1}{l}{4} & \multicolumn{1}{l}{-} & \multicolumn{1}{l}{-} & \multicolumn{1}{l}{$\mathbb{S}_i \times N$} \\
\hline
Zero point & 8 & $z$ & $z_i \times N$ & $z$ \\
\hline
\end{tabular}
\end{table}

\begin{algorithm}
\SetAlgoLined
input and its zero point: $q_{in}, z_{in}$ \\
output's zero point: $z_{out}$ \\
weights and its zero point: $q_w, z_w$ \\
bias: $q_B$ \\
scale compound of $\{s_{in}, s_w, s_{out}\} \rightarrow  M, s$ \\
ch-wise shift scale: $\mathbb{S}_i$ \\
\For{$0 \leq x < w_{out}, 0 \leq y < h_{out}, 0 \leq z < c_{out}$}{
    $acc_{32} = 0$ \\
    $ix\_ = x*stride - pad_{left}$ \\
    $iy\_ = y*stride - pad_{top}$ \\
    \For{$0 \leq i < w_w, 0 \leq j < h_w$}{
        $ix = ix\_ + i$ \\
        $iy = iy\_ + j$ \\
        \If{$0 \leq ix < w_{in}, 0 \leq iy < h_{in} $}{
            \For{$0 \leq k < c_{in}$}{
                $acc_{32} \mathrel{{+}{=}} (int32)(((int16)q_{in}[iy*w_{in}*c_{in} + ix*c_{in} + iz] - (int16)z_{in}) * ((int16)q_w[j*w_{w}*c_{in}*c_{out} + i*c_{in}*c_{out} + k*c_{out} + z] - (int16)z_{w}))$ \\
            }
        } 
    }
    $acc_{32} \mathrel{{+}{=}} q_B[z]$ \\
    $\bm{out_{32} =  M * (((s > 0)?(out_{32} \ll s):(out_{32} \gg -s)) \gg \mathbb{S}_{i=z}})$ \\
    $out_{32} = (relu) ? max(0, out_{32}) : out_{32}$ \\
    $out_{32} \mathrel{{+}{=}} z_{out}$ \\
    $out_{8}[y*w_{out}*c_{out} + x*c_{out} + z] = sat\_uint8(out_{32})$
}
\caption{WES-coupled fixed-point computing operator of Conv + ReLU (or ReLUN)}
\label{alg:fixed-computing}
\end{algorithm}

Herein, we can incorporate an inversion of channel-wise shift scale, $\mathbb{S}_i^{-1}$, by simply shifting the binary floating-point number by $\mathbb{S}_i$ to the right direction (Line 21 in Algorithm \ref{alg:fixed-computing}). Based on our experiments, 4 bits are enough to represent the candidates of channel-wise shift scales, $2^{\mathbb{Z}}$, $\mathbb{Z}\in\{0,...,15\}$. This binary shift is a key feature, which improves the accuracy performance without harming the latency due to a hardware-friendly bit-wise operation. The major scale compound and zero point are shared across channels, and furthermore the shift scale adjusts the range per channel, which is very cost-effective. Table \ref{tab:lwq_cwq_wes} shows the difference of quantization parameter specification among LWQ, CWQ, and WES. 

\section{Experimental Results}
We validate our proposed WES quantization method using up-to-date models in varying tasks compared to previous methods. We quantize only open models that already completed training.

\subsection{Baseline Methods}
Our interest is confined to uniform affine (with a zero point) quantization, which is known for its hardware-friendliness and fundamental needs. As representative methods to tackle the accuracy degradation after post-training quantization, channel-wise quantization (CWQ), and data-free quantization (DFQ) are the target baseline methods that our WES is compared to. Besides, layer-wise quantization (LWQ) is also included for the ablation study of WES.

\subsection{Experimental Setup}
\label{subsec:experimental_setup}
The computing environment is Python 3.7, Tensorflow (TF) 2.0 with a single GPU P40. The pre-trained models that are officially referred to in TF or are popularly cited are downloaded and we do not fine-tune the model. All convolutional layers including a depth-wise convolution and a fully-connected layer are quantized. For activation quantization, 1K representative images are selected at random in a train dataset but the label information is not used. Also, we do not make use of validation accuracy estimation in a cost or loss function. The result using DFQ was reproduced in our TF code migrated from PyTorch code\footnote[2]{https://github.com/jakc4103/DFQ}. For fair comparison, the activation ranges were safely set by feeding representative data, not predicting ranges from the batch normalization parameters ($\beta, \gamma$) proposed in DFQ.

\textbf{Weight clipping}: In uniform (or linear) quantization, quantization error is bounded by a half of the quantization step or scale which increments as the dynamic range increases \cite{gibson1998digital}. Assuming that weights can lie on the long-tailed distribution, the weight values outside a fixed interval are clipped to the interval edges. This leads to an increase in clipping error but a decrease in quantization error, so it is important to find the optimal range to make a balance between the errors. This clipping process is performed after the WES process just before layer-wise quantization. The reason is that some depth-wise weights in MobileNet before WES processing tend to be congregated in the narrow and distant range (Fig. \ref{fig:mv1_dist_before_after_wes}) in specific channels. To disperse the clipping evenly to other channels, it is recommended to apply the weight clipping after WES rescales weights. To find the optimal clipping range per layer, we use a heuristic non-linear optimization method (Nelder–Mead) where the cost is defined as the sum of clipping error and quantization error.

\textbf{Weight pruning}: To take advantage of weight sparsification, the pruning technique \cite{zhu2017prune} can be applied prior to the WES process. In NPU, this can help reduce DRAM bandwidth and SRAM size, thereby increasing power efficiency (FPS/W) without significantly compromising accuracy. To prune weights, the magnitude threshold is empirically set up. Then, unimportant elements, which have the magnitude of the weight value less than the threshold, are nullified to zero. The sparse weights are compressed by representing them as 1-bit masks with the same dimension of the original weights, compactly packed non-zero weights, and a single integer indicating the size of the compressed data.

\textbf{Activation quantization}: For activation quantization, layer-wise quantization is applied. Variants of convolution operator (e.g., normal, depthwise) and followed piece-wise linear activation function (e.g., ReLU, ReLU6) are fused for speeding-up inference and reducing fixed-computing error caused by the scaling and the datatype conversion. The dynamic range $[min, max]$ after activation function is obtained in the statistical way. In specific, we obtain min's and max's distributions by feeding representative images without labels to the network model and then the final $[min, max]$ is determined as the left and right clipping edges corresponding to the fixed percentile ($1\%$) in each distribution, respectively.

\subsection{Applications}
\textbf{Image classification:}
The ImageNet Large Scale Visual Recognition Challenge (ILSVRC) 2012 \cite{deng2009imagenet,russakovsky2015imagenet} is a large visual database, containing 1K sub-classes, 1.2M images as training dataset and 50K images as evaluation dataset. We choose various backbone models targeting ImageNet classification from mobile-oriented light models to cloud-oriented heavy models such as MobileNet v1/v2 \cite{howard2017mobilenets,sandler2018mobilenetv2}, EfficientNet B0/B3 \cite{tan2019efficientnet}, ResNet50 \cite{he2016deep}, and Inception v3 \cite{szegedy2016rethinking}. Only when applying DFQ to the MobileNet family, ReLU6 is replaced with ReLU due to the algorithm constraint, which is then finetuned for 40 epochs with 1e-5 initial learning rate. Otherwise, original pre-trained models are quantized. 

CIFAR-10 \cite{torralba200880} is a tiny image dataset that is a collection of 32x32 colour images in 10 classes, with 6K images per class. We employ ResNet56 that is a model trained on CIFAR-10 dataset.

\textbf{Multiple object detection:}
Localization and identification of multiple objects in images is a more advanced topic than image classification. Single shot multibox detection (SSD) \cite{liu2016ssd} eliminates miscellaneous processes and encapsulates all computation in a single network. Owing to these features, SSD is known for being faster and more accurate than previous YOLO \cite{redmon2016you} or Faster R-CNN \cite{ren2015faster} networks. Replacing the backbone network, VGG16, of SSD with effective MobileNet v1 invented the state-of-the-art modern object detection system, MobileNet-SSD \cite{howard2017mobilenets}, which is pre-trained on the Widerface dataset \cite{yang2016wider} for the face detection benchmark.

\textbf{Super resolution:}
Image super-resolution (SR) has been focused in the image processing area since the advent of deep learning. Fast Super-Resolution Convolutional Neural Networks (FSRCNN) \cite{dong2016accelerating} is one of well-known SR networks achieving both of accuracy and speed. The model is pre-trained on 91-images and is tested on Set14.

\subsection{Ablation Study}

In this section we explore the effect of our WES method on the image classification task. It is validated on MobileNet v1/v2 and EfficientNet B3 using ImageNet validation dataset. We assume the accuracy degradation arises from varying ranged and small overlapped distribution of weights across channels. To measure the difference and effect of the overlapping of LWQ and WES, a new metric, a so-called overlap ratio, is defined as follows: 
\begin{equation}
\text{Overlap ratio}=\frac{1}{N}\sum_{i=1}^{N} \frac{max_i-min_i}{max-min},
\end{equation}
where $[min_i, max_i]$ is a weight range of \textit{i}-th channel and $[min, max]$ is a range of all $N$ channels, i.e. the minimum of $min_i$ and the maximum of $max_i$. Besides, quantization error is measured in the fake quantization procedure and test accuracy is estimated using test dataset after UINT8 weight and activation quantizations. 

\begin{figure}[h]
\begin{subfigure}{.5\textwidth}
    \includegraphics[width=1\linewidth]{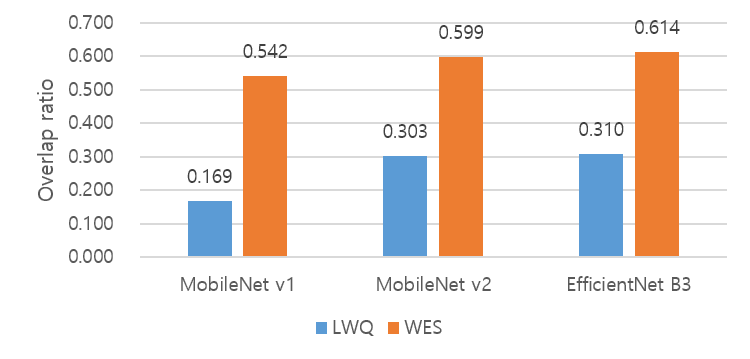} 
      \caption{Overlap ratio}
  \label{fig:sfig1}
\end{subfigure}
\begin{subfigure}{.5\textwidth}
    \includegraphics[width=1\linewidth]{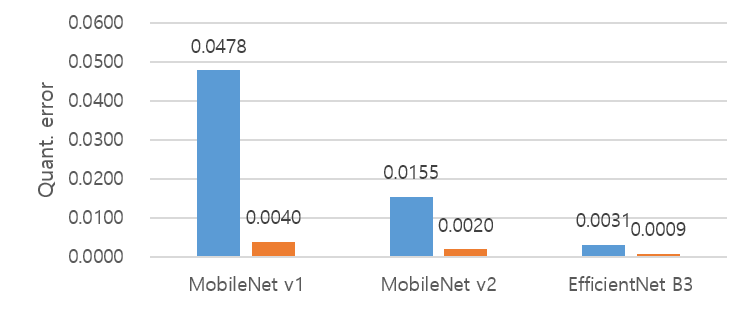} 
       \caption{Quantization error}
  \label{fig:sfig2}
\end{subfigure}   
\begin{subfigure}{.5\textwidth}
    \includegraphics[width=1\linewidth]{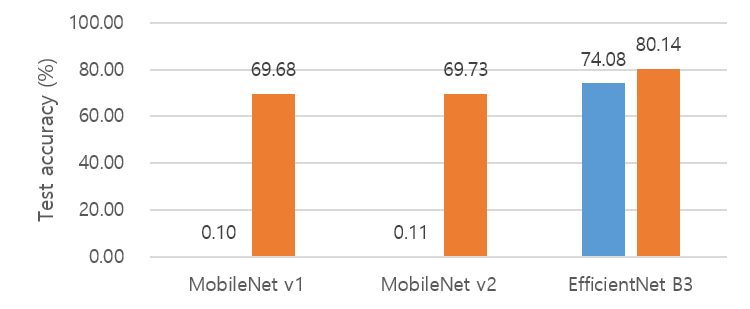} 
       \caption{Test accuracy ($\%$)}
  \label{fig:sfig3}
\end{subfigure}
\caption{Ablation study of WES (orange) compared to LWQ (blue) in perspectives of overlap ratio, quantization error, and test accuracy.}
\label{fig:ablation_study}
\end{figure}

As illustrated in Fig. \ref{fig:ablation_study}, WES shows significant improvement ($2\sim3\times$) in the overlap ratio over the baseline LWQ. The improvement is observed evenly in all the depth-wise and normal convolution layers. In the MobileNet and EfficientNet families including MBConv \cite{sandler2018mobilenetv2,tan2019efficientnet} block, the effect of WES is outstanding. Accordingly, the quantization error is reduced as much as $3\sim12\times$, which results in the accuracy recovery close to the FP32 base model ($70.60\%$ vs. $69.68\%$, $71.87\%$ vs. $69.73\%$, $81.38\%$ vs. $80.14\%$). This ablation study proves the success of WES in that independent weight shift scaling per channel is straightforward but very effective in greatly improving the precision of per-layer quantization.    
We also experimented that WES is applied to only depth-wise convolution layers while applying LWQ to normal convolution layers. This is because we considered reusing the most of existing LWQ-based fixed-computing operators but the depth-wise convolutions. The partial application of WES to the depth-wise convolutions yields the insufficient accuracy gain, $69.68\% \rightarrow 63.16\%$ in MobileNet v1. Thus, overall improvement of the range overlap in all convolution layers is important to achieve the best accuracy.

The iterative search was introduced for finding an optimal total range in Algorithm \ref{alg:WES}. The initial value of the variable is determined with a maximum value over all channel-wise ranges (deterministic). The iterative search based on the Nelder-Mead heuristic non-linear optimization further refines the variable (iterative). In order to confirm the effect of iterative search, we compare the performance resulted from the iterative search versus the deterministic way. Table \ref{tab:iterative_deterministic_wes} shows that the iterative search helps further improve the overlap ratio, reduce the quantization error, and increase the test accuracy ($69.5\%\rightarrow69.7\%$) in MobileNet v1. Note that if the quantization error is small enough, albeit that there exists slight improvement in quantization error, it always does not guarantee better test accuracy like MobileNet v2. 

\begin{table}[h]
\caption{Deterministic or iterative search of an optimal total range in MobileNet v1/v2. We report the test accuracy just with weight quantization to exclude the random effect by data shuffling in activation quantization.}
\label{tab:iterative_deterministic_wes}

\centering
\begin{tabular}{p{0.28\linewidth}p{0.15\linewidth}p{0.15\linewidth}p{0.15\linewidth}}
\hline
  & Overlap Ratio & Quant. Error & Test Acc. ($\%$) \\
\hline
\hline
 MobileNet v1 &  & & \\
\hline
Deterministic & 0.535 & 0.00402 & 69.5 \\
Iterative & 0.542 & 0.00398 & 69.7 \\
\hline
 MobileNet v2 &  & & \\
\hline
Deterministic & 0.586 & 0.00202 & 69.8 \\
Iterative & 0.599 & 0.00199 & 69.8 \\
\hline
\end{tabular}
\end{table}

%

\subsection{Comparison to Other Methods and Applications}
 The performance of WES is compared with LWQ, DFQ \cite{nagel2019data}, and CWQ in various application domains including image classification, object detection and super-resolution. Table \ref{tab:post_train_quant_accuracy} summarizes the post-training quantization results in UINT8. For MobileNet v1/v2, the results applying weight clipping with a fixed percentile are also shown.
 
 \begin{table*}[ht]
\caption{Accuracy of post-training quantization (UINT8). The symbol ${}^{\mathbf{wc}}$ denotes the result from additional application of weight clipping. The clipping range is optimized using a heuristic non-linear optimization method (See the details in the section of Experimental Setup).}
\label{tab:post_train_quant_accuracy}
\centering
\begin{tabular}{p{0.09\linewidth}p{0.11\linewidth}p{0.07\linewidth}p{0.07\linewidth}p{0.07\linewidth}p{0.07\linewidth}p{0.07\linewidth}p{0.07\linewidth}}
\hline
Dataset & Model & Accuracy & Baseline (FP32) & LWQ & CWQ & DFQ & WES \\
\hline
\hline
    \multirow{5}{*}{ImageNet} 
                              & \multicolumn{1}{l}{MobileNet v1} & Top1($\%$) & \multicolumn{1}{l}{$70.60$} & \multicolumn{1}{l}{$0.10$} &                        \multicolumn{1}{l}{$69.74$ / $\mathbf{69.97^{wc}}$} & \multicolumn{1}{l}{$66.67$ / $67.52^{wc}$} & \multicolumn{1}{l}{$69.68$ / $69.78^{wc}$} \\
                              & \multicolumn{1}{l}{MobileNet v2} & Top1($\%$) & \multicolumn{1}{l}{$71.87$} & \multicolumn{1}{l}{$0.11$} &                        \multicolumn{1}{l}{$70.98$ / $\mathbf{70.98^{wc}}$} & \multicolumn{1}{l}{$69.12$} & \multicolumn{1}{l}{$69.73$ / $70.96^{wc}$} \\
                              & \multicolumn{1}{l}{EfficientNet B3} & Top1($\%$) & \multicolumn{1}{l}{$81.38$} & \multicolumn{1}{l}{$74.08$} &                        \multicolumn{1}{l}{$\mathbf{80.20}$} & \multicolumn{1}{l}{-} & \multicolumn{1}{l}{80.14} \\
                              & \multicolumn{1}{l}{ResNet50 v1} & Top1($\%$) & \multicolumn{1}{l}{$76.48$} & \multicolumn{1}{l}{$76.18$} &                                                  \multicolumn{1}{l}{$76.20$} & \multicolumn{1}{l}{-} &                                                                    \multicolumn{1}{l}{$\mathbf{76.29}$} \\
                              & \multicolumn{1}{l}{Inception v3} & Top1($\%$) & \multicolumn{1}{l}{$77.97$} & \multicolumn{1}{l}{$77.62$} &                        \multicolumn{1}{l}{$77.61$} & \multicolumn{1}{l}{-} & \multicolumn{1}{l}{$\mathbf{77.64}$} \\
\hline
Cifar10 & ResNet56  & Top1($\%$)  & $93.37$ & $93.28$ & $93.18$ & - & $\mathbf{93.29}$ \\
\hline
Widerface & MobileNet-SSD & mAP($\%$) & $54.29$ & $52.32$ & $\mathbf{55.20}$ & - & $55.05$ \\
\hline
91Im/Set14 & FSRCNN & PSNR(dB) & $30.79$ & $30.62$ & $30.59$ & - & $\mathbf{30.64}$ \\
                &       & SSIM & $0.898$ & $0.893$ & $0.893$ & - & $\mathbf{0.894}$ \\

\hline
\end{tabular}
\end{table*}

 In detail, WES outperforms DFQ for MobileNet v1 ($69.78\%$ vs. $67.52\%$) and MobileNet v2 ($70.96\%$ vs. $69.12\%$) in accuracy. DFQ employs bias correction in addition to cross-layer equalization so as to correct for the biased quantization error in the bias. Nevertheless, our reproduced results are similar to or worse than ones without bias correction.  
 
 In general, it is shown that WES is very competitive to CWQ in all the target applications in this paper. CWQ shows slightly better performance over WES in MobileNet v1 ($69.97\%$ vs. $69.78\%)$, MobileNet v2 ($70.98\%$ vs. $70.96\%$), and EfficientNet ($80.20\%$ vs. $80.14\%$) containing depth-wise convolution layers. However, WES gives better or similar performance in the other models -- ResNet50 v1 ($76.29\%$ vs. $76.20\%$), Inception v3 ($77.64\%$ vs. $77.61\%$), ResNet56 ($93.29\%$ vs. $93.18\%$), FSRCNN ($30.64$dB vs. $30.59$dB) than CWQ. Carefully speaking, based on our experiments, the proposed WES method proves its robust performance in varying applications.

\subsection{Memory Cost}
\begin{figure}[h!]
\centering
    \includegraphics[width=0.5\textwidth]{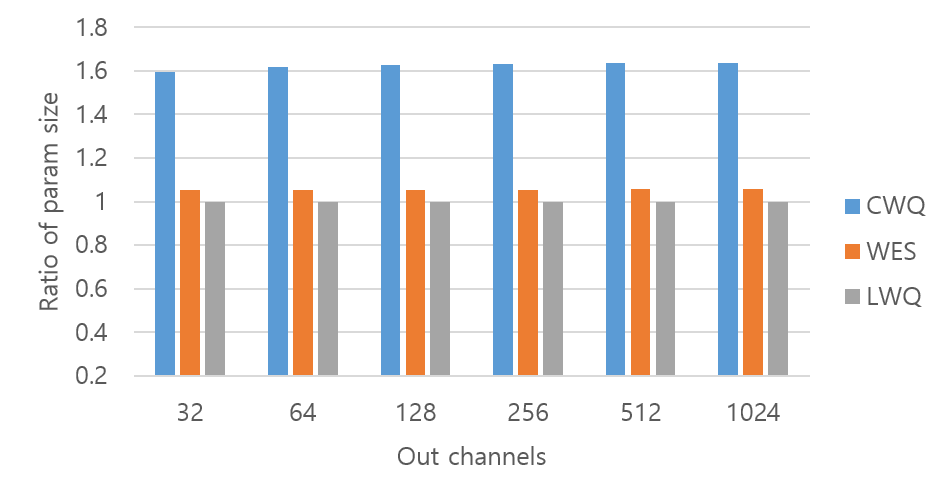}
    (a)
    \includegraphics[width=0.5\textwidth]{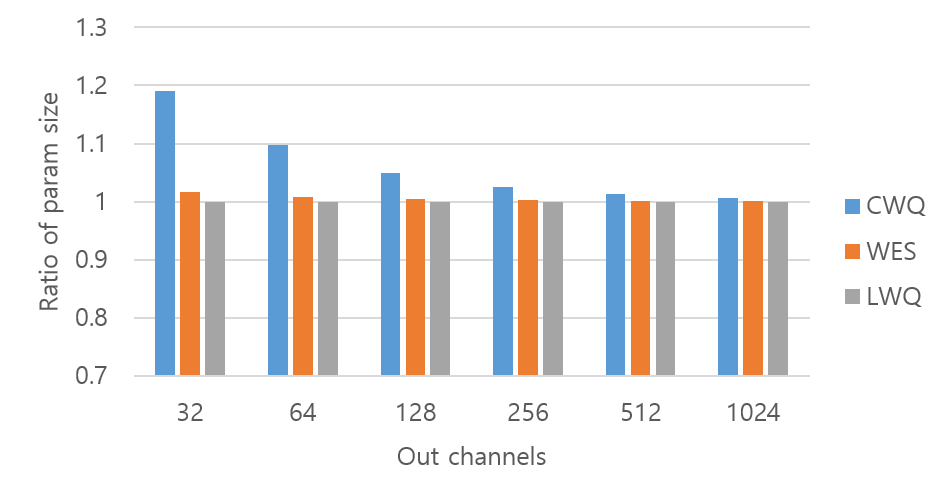}
    (b)
    \caption{Quantized parameter size of (a) 3x3 depth-wise convolution and (b) 1x1 point-wise convolution for LWQ, CWQ, and WES for different number of channels. A sparsity of $20\%$ by weight pruning is assumed in 1x1 point-wise convolution.}
    \label{fig:param_comp}
\end{figure}

With respect to the size of quantized parameters, WES needs just extra 4 bits for storing shift scale factor for each channel, on the other hand, CWQ needs a (32+6)-bit scale compound, a 8-bit zero point of weights for each channel. As shown in Fig. \ref{fig:param_comp} (a), in 3x3 depth-wise convolution, WES slightly increases by $5.5\%$, but CWQ increases by $59\% \sim 64\%$ compared to LWQ consistently over the number of channels. By contrast, in normal convolution, the difference of CWQ or WES against LWQ in size is negligible because kernel weights occupy much larger portion out of the overall than depth-wise convolution. Table \ref{tab:post_train_quant_params} shows the overall size of naive quantized parameters without weight compression. For CWQ, parameter increment of around $3\%$ incurs but for WES around $0.2\%$. 

If we further compress pruned weights into a sparse format, the advantage of WES saving the quantization parameters is clearly present in normal convolution as Fig. \ref{fig:param_comp} (b). The benefit of WES is shown differently depending on the number of channels. For example, applying a sparsity of $20\%$ by weight compression is shown to preserve the acccuracy within $1\%$ drop. The parameter size increases by $0.1\%$ to $1.7\%$ for 32 to 1024 channels in WES but by $0.6\%$ to $19.1\%$ in CWQ compared to LWQ. As the sparsity gets bigger, the effect will be larger.

\begin{table*}[t]
\caption{Quantized parameter size (MB) of post-training quantization (UINT8) without weight compression}
\label{tab:post_train_quant_params}
\centering
\begin{tabular}{p{0.10\linewidth}p{0.13\linewidth}p{0.08\linewidth}p{0.13\linewidth}p{0.10\linewidth}p{0.10\linewidth}}
\hline
Dataset & Model & Baseline (FP32) & LWQ / DFQ & CWQ & WES \\
\hline
\hline
    \multirow{5}{*}{ImageNet} 
                              & \multicolumn{1}{l}{MobileNet v1} & \multicolumn{1}{l}{16.5} & \multicolumn{1}{l}{4.16} &                        \multicolumn{1}{l}{4.22} & \multicolumn{1}{l}{4.16} \\
                              & \multicolumn{1}{l}{MobileNet v2} & \multicolumn{1}{l}{13.7} & \multicolumn{1}{l}{3.46} &                        \multicolumn{1}{l}{3.56} & \multicolumn{1}{l}{3.47} \\
                              & \multicolumn{1}{l}{EfficientNet B3} & \multicolumn{1}{l}{47.8} & \multicolumn{1}{l}{12.10} &                        \multicolumn{1}{l}{12.46} & \multicolumn{1}{l}{12.13} \\
                              & \multicolumn{1}{l}{ResNet50 v1} & \multicolumn{1}{l}{99.8} & \multicolumn{1}{l}{25.0} &                             \multicolumn{1}{l}{25.2} & \multicolumn{1}{l}{25.0} \\
                              & \multicolumn{1}{l}{Inception v3} & \multicolumn{1}{l}{93.0} & \multicolumn{1}{l}{23.3} &                        \multicolumn{1}{l}{23.4} & \multicolumn{1}{l}{23.3} \\

\hline
\end{tabular}
\end{table*}

\section{Conclusion}
We  addressed  the  accuracy  degradation  problem  after  quantizing weights of a varying distribution per channel in layer-wise quantization. The proposed method introduced a new weight equalizing shift scaler (WES)-coupled post training quantization having a new quantized  parameter called “channel-wise  shift scale” and its inverse scaling’s fusion to a convolutional operator in the neural processing unit (NPU). This method significantly improved the accuracy performance without harming the memory footprint. The ablation and experimental study showed that WES can successfully resolve the accuracy degradation problem of layer-wise quantization. In addition, its robust performance was proved over varying applications. 

\section{Future Works}
As future works, we will measure the system cycle for WES and baseline methods in the simulator environment of our custom-designed NPU. For automating the decision of the optimal pruning percentile for weight compression, optimization-based method will be tried. Other than vision tasks, we will expand validation to language tasks like acoustic speech recognition or natural language understanding.

\section{Acknowledgement}
This work is supported by Samsung Research, Samsung Electronics Co.Ltd.


\bibstyle{aaai21.bst}
\bibliography{quant.bib}

\begin{thebibliography}{36}
\providecommand{\natexlab}[1]{#1}
\providecommand{\url}[1]{\texttt{#1}}
\providecommand{\urlprefix}{URL }
\expandafter\ifx\csname urlstyle\endcsname\relax
  \providecommand{\doi}[1]{doi:\discretionary{}{}{}#1}\else
  \providecommand{\doi}{doi:\discretionary{}{}{}\begingroup
  \urlstyle{rm}\Url}\fi

\bibitem[{Albericio et~al.(2017)Albericio, Delm{\'a}s, Judd, Sharify, O'Leary,
  Genov, and Moshovos}]{albericio2017bit}
Albericio, J.; Delm{\'a}s, A.; Judd, P.; Sharify, S.; O'Leary, G.; Genov, R.;
  and Moshovos, A. 2017.
\newblock Bit-pragmatic deep neural network computing.
\newblock In \emph{Proceedings of the 50th Annual IEEE/ACM International
  Symposium on Microarchitecture}, 382--394.

\bibitem[{Banner, Nahshan, and Soudry(2019)}]{banner2019post}
Banner, R.; Nahshan, Y.; and Soudry, D. 2019.
\newblock Post training 4-bit quantization of convolutional networks for
  rapid-deployment.
\newblock In \emph{Advances in Neural Information Processing Systems},
  7950--7958.

\bibitem[{Cai et~al.(2020)Cai, Yao, Dong, Gholami, Mahoney, and
  Keutzer}]{cai2020zeroq}
Cai, Y.; Yao, Z.; Dong, Z.; Gholami, A.; Mahoney, M.~W.; and Keutzer, K. 2020.
\newblock Zeroq: A novel zero shot quantization framework.
\newblock In \emph{Proceedings of the IEEE/CVF Conference on Computer Vision
  and Pattern Recognition}, 13169--13178.

\bibitem[{Chen, Emer, and Sze(2016)}]{chen2016eyeriss}
Chen, Y.-H.; Emer, J.; and Sze, V. 2016.
\newblock Eyeriss: A spatial architecture for energy-efficient dataflow for
  convolutional neural networks.
\newblock \emph{ACM SIGARCH Computer Architecture News} 44(3): 367--379.

\bibitem[{Courbariaux, Bengio, and David(2015)}]{courbariaux2015binaryconnect}
Courbariaux, M.; Bengio, Y.; and David, J.-P. 2015.
\newblock Binaryconnect: Training deep neural networks with binary weights
  during propagations.
\newblock In \emph{Advances in neural information processing systems},
  3123--3131.

\bibitem[{Deng et~al.(2009)Deng, Dong, Socher, Li, Li, and
  Fei-Fei}]{deng2009imagenet}
Deng, J.; Dong, W.; Socher, R.; Li, L.-J.; Li, K.; and Fei-Fei, L. 2009.
\newblock Imagenet: A large-scale hierarchical image database.
\newblock In \emph{2009 IEEE conference on computer vision and pattern
  recognition}, 248--255. Ieee.

\bibitem[{Dong, Loy, and Tang(2016)}]{dong2016accelerating}
Dong, C.; Loy, C.~C.; and Tang, X. 2016.
\newblock Accelerating the super-resolution convolutional neural network.
\newblock In \emph{European conference on computer vision}, 391--407. Springer.

\bibitem[{Gibson et~al.(1998)Gibson, Berger, Lookabaugh, Baker, and
  Lindbergh}]{gibson1998digital}
Gibson, J.~D.; Berger, T.; Lookabaugh, T.; Baker, R.; and Lindbergh, D. 1998.
\newblock \emph{Digital compression for multimedia: principles and standards}.
\newblock Morgan Kaufmann.

\bibitem[{Han, Mao, and Dally(2015)}]{han2015deep}
Han, S.; Mao, H.; and Dally, W.~J. 2015.
\newblock Deep compression: Compressing deep neural networks with pruning,
  trained quantization and huffman coding.
\newblock \emph{arXiv preprint arXiv:1510.00149} .

\bibitem[{He et~al.(2016)He, Zhang, Ren, and Sun}]{he2016deep}
He, K.; Zhang, X.; Ren, S.; and Sun, J. 2016.
\newblock Deep residual learning for image recognition.
\newblock In \emph{Proceedings of the IEEE conference on computer vision and
  pattern recognition}, 770--778.

\bibitem[{Horowitz(2014)}]{horowitz20141}
Horowitz, M. 2014.
\newblock 1.1 computing's energy problem (and what we can do about it).
\newblock In \emph{2014 IEEE International Solid-State Circuits Conference
  Digest of Technical Papers (ISSCC)}, 10--14. IEEE.

\bibitem[{Howard et~al.(2017)Howard, Zhu, Chen, Kalenichenko, Wang, Weyand,
  Andreetto, and Adam}]{howard2017mobilenets}
Howard, A.~G.; Zhu, M.; Chen, B.; Kalenichenko, D.; Wang, W.; Weyand, T.;
  Andreetto, M.; and Adam, H. 2017.
\newblock Mobilenets: Efficient convolutional neural networks for mobile vision
  applications.
\newblock \emph{arXiv preprint arXiv:1704.04861} .

\bibitem[{Hubara et~al.(2016)Hubara, Courbariaux, Soudry, El-Yaniv, and
  Bengio}]{hubara2016binarized}
Hubara, I.; Courbariaux, M.; Soudry, D.; El-Yaniv, R.; and Bengio, Y. 2016.
\newblock Binarized neural networks.
\newblock In \emph{Advances in neural information processing systems},
  4107--4115.

\bibitem[{Jacob et~al.(2018)Jacob, Kligys, Chen, Zhu, Tang, Howard, Adam, and
  Kalenichenko}]{jacob2018quantization}
Jacob, B.; Kligys, S.; Chen, B.; Zhu, M.; Tang, M.; Howard, A.; Adam, H.; and
  Kalenichenko, D. 2018.
\newblock Quantization and training of neural networks for efficient
  integer-arithmetic-only inference.
\newblock In \emph{Proceedings of the IEEE Conference on Computer Vision and
  Pattern Recognition}, 2704--2713.

\bibitem[{Jouppi et~al.(2017)Jouppi, Young, Patil, Patterson, Agrawal, Bajwa,
  Bates, Bhatia, Boden, Borchers et~al.}]{jouppi2017datacenter}
Jouppi, N.~P.; Young, C.; Patil, N.; Patterson, D.; Agrawal, G.; Bajwa, R.;
  Bates, S.; Bhatia, S.; Boden, N.; Borchers, A.; et~al. 2017.
\newblock In-datacenter performance analysis of a tensor processing unit.
\newblock In \emph{Proceedings of the 44th Annual International Symposium on
  Computer Architecture}, 1--12.

\bibitem[{Judd et~al.(2016)Judd, Albericio, Hetherington, Aamodt, and
  Moshovos}]{judd2016stripes}
Judd, P.; Albericio, J.; Hetherington, T.; Aamodt, T.~M.; and Moshovos, A.
  2016.
\newblock Stripes: Bit-serial deep neural network computing.
\newblock In \emph{2016 49th Annual IEEE/ACM International Symposium on
  Microarchitecture (MICRO)}, 1--12. IEEE.

\bibitem[{Jung et~al.(2019)Jung, Son, Lee, Son, Han, Kwak, Hwang, and
  Choi}]{jung2019learning}
Jung, S.; Son, C.; Lee, S.; Son, J.; Han, J.-J.; Kwak, Y.; Hwang, S.~J.; and
  Choi, C. 2019.
\newblock Learning to quantize deep networks by optimizing quantization
  intervals with task loss.
\newblock In \emph{Proceedings of the IEEE Conference on Computer Vision and
  Pattern Recognition}, 4350--4359.

\bibitem[{Krishnamoorthi(2018)}]{krishnamoorthi2018quantizing}
Krishnamoorthi, R. 2018.
\newblock Quantizing deep convolutional networks for efficient inference: A
  whitepaper.
\newblock \emph{arXiv preprint arXiv:1806.08342} .

\bibitem[{Lee, Kapoor, and Kim(2018)}]{lee2018deeptwist}
Lee, D.; Kapoor, P.; and Kim, B. 2018.
\newblock Deeptwist: Learning model compression via occasional weight
  distortion.
\newblock \emph{arXiv preprint arXiv:1810.12823} .

\bibitem[{Lee et~al.(2018)Lee, Kim, Kang, Shin, Kim, and Yoo}]{lee2018unpu}
Lee, J.; Kim, C.; Kang, S.; Shin, D.; Kim, S.; and Yoo, H.-J. 2018.
\newblock UNPU: A 50.6 TOPS/W unified deep neural network accelerator with
  1b-to-16b fully-variable weight bit-precision.
\newblock In \emph{2018 IEEE International Solid-State Circuits
  Conference-(ISSCC)}, 218--220. IEEE.

\bibitem[{Liu et~al.(2016)Liu, Anguelov, Erhan, Szegedy, Reed, Fu, and
  Berg}]{liu2016ssd}
Liu, W.; Anguelov, D.; Erhan, D.; Szegedy, C.; Reed, S.; Fu, C.-Y.; and Berg,
  A.~C. 2016.
\newblock Ssd: Single shot multibox detector.
\newblock In \emph{European conference on computer vision}, 21--37. Springer.

\bibitem[{Liu et~al.(2020)Liu, Ye, Zhou, and Liu}]{liu2020post}
Liu, X.; Ye, M.; Zhou, D.; and Liu, Q. 2020.
\newblock Post-training Quantization with Multiple Points: Mixed Precision
  without Mixed Precision.
\newblock \emph{arXiv preprint arXiv:2002.09049} .

\bibitem[{Nagel et~al.(2019)Nagel, Baalen, Blankevoort, and
  Welling}]{nagel2019data}
Nagel, M.; Baalen, M.~v.; Blankevoort, T.; and Welling, M. 2019.
\newblock Data-free quantization through weight equalization and bias
  correction.
\newblock In \emph{Proceedings of the IEEE International Conference on Computer
  Vision}, 1325--1334.

\bibitem[{Nelder and Mead(1965)}]{nelder1965simplex}
Nelder, J.~A.; and Mead, R. 1965.
\newblock A simplex method for function minimization.
\newblock \emph{The computer journal} 7(4): 308--313.

\bibitem[{Pelikan et~al.(1999)Pelikan, Goldberg, Cant{\'u}-Paz
  et~al.}]{pelikan1999boa}
Pelikan, M.; Goldberg, D.~E.; Cant{\'u}-Paz, E.; et~al. 1999.
\newblock BOA: The Bayesian optimization algorithm.
\newblock In \emph{Proceedings of the genetic and evolutionary computation
  conference GECCO-99}, volume~1, 525--532. Citeseer.

\bibitem[{Redmon et~al.(2016)Redmon, Divvala, Girshick, and
  Farhadi}]{redmon2016you}
Redmon, J.; Divvala, S.; Girshick, R.; and Farhadi, A. 2016.
\newblock You only look once: Unified, real-time object detection.
\newblock In \emph{Proceedings of the IEEE conference on computer vision and
  pattern recognition}, 779--788.

\bibitem[{Ren et~al.(2015)Ren, He, Girshick, and Sun}]{ren2015faster}
Ren, S.; He, K.; Girshick, R.; and Sun, J. 2015.
\newblock Faster r-cnn: Towards real-time object detection with region proposal
  networks.
\newblock In \emph{Advances in neural information processing systems}, 91--99.

\bibitem[{Russakovsky et~al.(2015)Russakovsky, Deng, Su, Krause, Satheesh, Ma,
  Huang, Karpathy, Khosla, Bernstein et~al.}]{russakovsky2015imagenet}
Russakovsky, O.; Deng, J.; Su, H.; Krause, J.; Satheesh, S.; Ma, S.; Huang, Z.;
  Karpathy, A.; Khosla, A.; Bernstein, M.; et~al. 2015.
\newblock Imagenet large scale visual recognition challenge.
\newblock \emph{International journal of computer vision} 115(3): 211--252.

\bibitem[{Sandler et~al.(2018)Sandler, Howard, Zhu, Zhmoginov, and
  Chen}]{sandler2018mobilenetv2}
Sandler, M.; Howard, A.; Zhu, M.; Zhmoginov, A.; and Chen, L.-C. 2018.
\newblock Mobilenetv2: Inverted residuals and linear bottlenecks.
\newblock In \emph{Proceedings of the IEEE conference on computer vision and
  pattern recognition}, 4510--4520.

\bibitem[{Szegedy et~al.(2016)Szegedy, Vanhoucke, Ioffe, Shlens, and
  Wojna}]{szegedy2016rethinking}
Szegedy, C.; Vanhoucke, V.; Ioffe, S.; Shlens, J.; and Wojna, Z. 2016.
\newblock Rethinking the inception architecture for computer vision.
\newblock In \emph{Proceedings of the IEEE conference on computer vision and
  pattern recognition}, 2818--2826.

\bibitem[{Tan and Le(2019)}]{tan2019efficientnet}
Tan, M.; and Le, Q.~V. 2019.
\newblock Efficientnet: Rethinking model scaling for convolutional neural
  networks.
\newblock \emph{arXiv preprint arXiv:1905.11946} .

\bibitem[{Torralba, Fergus, and Freeman(2008)}]{torralba200880}
Torralba, A.; Fergus, R.; and Freeman, W.~T. 2008.
\newblock 80 million tiny images: A large data set for nonparametric object and
  scene recognition.
\newblock \emph{IEEE transactions on pattern analysis and machine intelligence}
  30(11): 1958--1970.

\bibitem[{Yang et~al.(2016)Yang, Luo, Loy, and Tang}]{yang2016wider}
Yang, S.; Luo, P.; Loy, C.~C.; and Tang, X. 2016.
\newblock WIDER FACE: A Face Detection Benchmark.
\newblock In \emph{IEEE Conference on Computer Vision and Pattern Recognition
  (CVPR)}.

\bibitem[{Zhao et~al.(2019{\natexlab{a}})Zhao, Hu, Dotzel, De~Sa, and
  Zhang}]{zhao2019improving}
Zhao, R.; Hu, Y.; Dotzel, J.; De~Sa, C.; and Zhang, Z. 2019{\natexlab{a}}.
\newblock Improving neural network quantization without retraining using
  outlier channel splitting.
\newblock \emph{arXiv preprint arXiv:1901.09504} .

\bibitem[{Zhao et~al.(2019{\natexlab{b}})Zhao, Wang, Cai, Liu, and
  Zhang}]{zhao2019linear}
Zhao, X.; Wang, Y.; Cai, X.; Liu, C.; and Zhang, L. 2019{\natexlab{b}}.
\newblock Linear Symmetric Quantization of Neural Networks for Low-precision
  Integer Hardware.
\newblock In \emph{International Conference on Learning Representations}.

\bibitem[{Zhu and Gupta(2017)}]{zhu2017prune}
Zhu, M.; and Gupta, S. 2017.
\newblock To prune, or not to prune: exploring the efficacy of pruning for
  model compression.
\newblock \emph{arXiv preprint arXiv:1710.01878} .

\end{thebibliography}

\end{document}